\documentclass{article}

\usepackage{arxiv}

\usepackage[utf8]{inputenc} 
\usepackage[T1]{fontenc}    
\usepackage{hyperref}       
\usepackage{url}            
\usepackage{booktabs}       
\usepackage{amsfonts}       

\usepackage{times}
\usepackage{epsfig}
\usepackage{graphicx}
\usepackage{amsmath}
\usepackage{amssymb}
\usepackage{multirow}

\title{Semi-Supervised Cross-Modal Retrieval with Label Prediction}

\author{
  Devraj ~Mandal\thanks{corresponding author} \\
  Electrical Engineering Dept.\\
  Indian Institute of Science\\
  Bangalore, India 560012 \\
  \texttt{devrajm@iisc.ac.in} \\
   \And
 Pramod ~Rao \thanks{work done while interning at IISc} \\
  Electrical Engineering Dept.\\
  Indian Institute of Science\\
  Bangalore, India 560012 \\
  \texttt{pramod$\_$rao@outlook.com} \\
    \And
 Soma ~Biswas \\
  Electrical Engineering Dept.\\
  Indian Institute of Science\\
  Bangalore, India 560012 \\
  \texttt{somabiswas@iisc.ac.in} \\
}

\begin{document}
\maketitle

\begin{abstract}

Due to abundance of data from multiple modalities, cross-modal retrieval tasks with image-text, audio-image, etc. are gaining increasing importance.
Of the different approaches proposed, supervised methods usually give significant improvement over their unsupervised counterparts at the additional cost of labeling or annotation of the training data.
Semi-supervised methods are recently becoming popular as they provide an elegant framework to balance the conflicting requirement of labeling cost and accuracy. 
In this work, we propose a novel deep semi-supervised
framework which can seamlessly handle both labeled as well as unlabeled data.
The network has two important components: (a) the label prediction component predicts the labels for the unlabeled portion of the data and then (b) a common modality-invariant representation is learned for cross-modal retrieval.
The two parts of the network are trained sequentially one after the other. 
Extensive experiments on three standard benchmark datasets, Wiki, Pascal VOC and NUS-WIDE demonstrate that the proposed framework outperforms the state-of-the-art for both supervised and semi-supervised settings.
\end{abstract}


\section{Introduction}

\begin{figure}[t!]
	\begin{center}		
		\includegraphics[width=10.4cm,height=5.5cm]{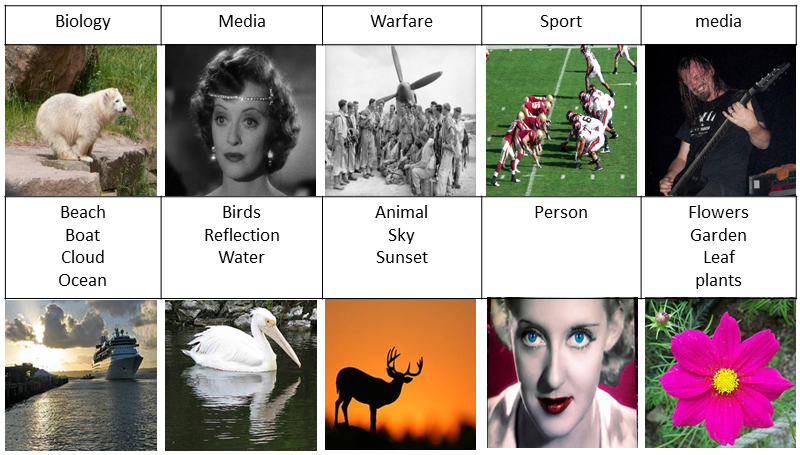}
	\end{center}
	\caption{Some illustrative examples of single and multi-label data from two datasets - Wiki \cite{wiki} (top row) and NUS-WIDE \cite{nus} (bottom row). Observe that the presence of multiple tags helps to define non-binary relationships between the example data.}
	\label{example_images}
	\vspace{- 10 pt}
\end{figure}

The steady increase in the amount of multimedia data like images, texts, sketches, audio, video, etc. over the last several years have made cross-modal retrieval a very active area of research.
As an example, given an image or sketch, we may want to retrieve the textual documents semantically related to it.
Though some of the data samples are associated with single labels (or tags)~\cite{wiki}, usually they can be better described using multiple labels~\cite{pascal} \cite{nus} since non-binary similarity between examples can be better captured by using multiple labels.
Few illustrative examples from different datasets are shown in Figure \ref{example_images}.

The cross-modal retrieval algorithms can be broadly classified under three categories, namely (1) unsupervised (b) supervised and (c) semi-supervised. 
Supervised algorithms usually outperform their unsupervised counterparts due to the presence of labeled training data, which is often quite expensive to obtain.
Semi-supervised frameworks \cite{gssl}\cite{jrl}\cite{s2upg} serve as a trade-off between these two important but conflicting requirements of performance and labeling cost.
This is achieved by utilising a small amount of labeled data and a large amount of unlabeled data (which is easy to obtain but costly to annotate).

In this work, we develop a deep learning framework for cross-modal retrieval which can seamlessly work under both supervised and semi-supervised settings.
The proposed framework consists of two main components, namely (a) a label prediction module and (b) a 
common representation learning module.
The first module is used to effectively predict the labels for the unlabeled portion of the training data, whereas a common modality-invariant representation corresponding to the input modalities is learned in the second module.
We propose different losses, so that both the labeled and unlabeled data can be utilized to learn an effective 
common representation for retrieving cross-modal data.
The two parts of the network are trained sequentially one after the other. 
Extensive experiments on three standard benchmark datasets, namely Wiki \cite{wiki}, Pascal VOC 2007 \cite{pascal} and NUS-WIDE \cite{nus} and comparisons with state-of-the-art cross-modal techniques show the effectiveness of the proposed approach. 
The main contributions of the proposed work can be summarized as follows
\begin{itemize}
	\item We propose a deep learning framework for cross-modal retrieval for semi-supervised setting.
	\item To the best of our knowledge, this is the first deep learning framework which has a label prediction framework for cross-modal retrieval to handle the unlabeled examples. 
	\item The proposed framework can seamlessly handle different scenarios - supervised and unsupervised settings,  single-label and multi-label data which are either paired or unpaired.
	\item  Extensive experiments show the effectiveness of the proposed approach.
	Specifically, in the challenging scenario when the amount of labeled training data is less, it significantly outperforms the state-of-the-art.
\end{itemize}


\section{Related Work}

Cross-modal retrieval is an active area of research and the different approaches proposed in literature can be divided into unsupervised, supervised and semi-supervised.
Unsupervised approaches do not have access to the training labels and in general utilizes the correspondence between the data of the two modalities to learn a common space.
Canonical Correlation Analysis (CCA) and its kernelized version (KCCA) \cite{cca} tries to project the data from the different modalities so that they become correlated.
Partial Least Squares~\cite{pls} linearly maps the different modalities into a common space. 
In~\cite{hwang}, connections between the objects in an image with keywords in the textual queries are established to design better cross-modal associations.

The supervised methods uses the class labels (single or multiple labels) to generate more meaningful connections between the data of the two modalities.
Deep CCA \cite{dcca} and Deep CCA AE \cite{dccae} integrate the concept of correlation learning into an encoder-decoder framework with reconstruction losses to learn the common space.
The discriminant latent space is learned by using the label information in the work of \cite{gma}.
Multiview Discriminant Analysis (MvDA) \cite{mvda} uses both the intra and inter domain relationships to capture more discriminative information for effective design of the common space. 
The work in \cite{wang} uses the concept of $l_{2,1}$ penalties on the projection functions for selecting discriminative and relevant features from the learned common domain.
Cluster CCA (CCCA) \cite{ccca} uses group class label information to effectively design the common space and its enhanced version multi-label CCA (ml-CCA) \cite{mlcca} can handle multi-label data.
Another algorithm which can effortlessly handle multi-label data to compute good common domain representation is the work in \cite{kang}.
Dictionary based supervised methods for cross-modal retrieval has been proposed in GCDL \cite{gcdl} and S2CDL \cite{s2cdl}.
CCCA \cite{ccca}, ml-CCCA \cite{mlcca} and GCDL \cite{gcdl} can handle both paired and unpaired data.
Deep supervised cross-modal algorithms have also been developed in the works of \cite{ngiam} and \cite{srivastava}.

Semi-supervised methods~\cite{jrl} \cite{s2upg} \cite{gssl} try to bridge the gap between the supervised and unsupervised methods by utilizing both labeled as well as unlabeled data for training, and it is much less explored compared to the others.
\cite{jrl} establishes an unified optimization framework to jointly associate the relationship between the correlation and the semantic information of the training examples. 
Regularization is used in \cite{jrl} to align the modalities more closely while making the model more robust to noise. 
In~\cite{s2upg}, a special patch graph regularization term is used to capture the underlying graphical structure of the different modalities jointly, which helps to integrate the complementary information.
GSS-SL \cite{gssl} captures the intrinsic manifold of the different modalities by using a label graph constraint and utilizes the label space for the common representation.
It uses a combination of label-linked loss and graph regularization to jointly predict the labels for the unlabeled data and learn the common space effectively.

The concept of label cleaning and prediction from weakly annotated or unlabeled data for single modality has been explored in~\cite{minsup} \cite{frenay}.
Algorithms like \cite{noise1} \cite{noise2} \cite{noise3} design noise-robust procedures and try to correct the mislabeled data.
Some approaches~\cite{noise4} \cite{noise5} \cite{minsup} select a small subset of the data whose clean annotations are available and tries to learn functions so as to clean the label noise.
The work in \cite{minsup} jointly learns to clean noisy annotations and design an effective image classifier.
Our work focuses on label prediction for cross-modal data and to the best of our knowledge is the first work in this field.

Another active research area closely related to cross-modal retrieval is cross-modal hashing~\cite{dcmh} \cite{var_hash}, where data from different modalities are projected into a discrete domain for fast and accurate retrieval. 
The cross-modal hashing techniques differ from the standard cross-modal retrieval in some important aspects, namely (a) projection into the discrete hamming space and (b) the retrieval data being the same as the training data.
In cross-modal retrieval, the retrieval set is completely disjoint from the training set, making the problem more challenging.



\section{The Proposed Approach}

Here, we describe in details the proposed deep learning framework for semi-supervised cross-modal retrieval.
The proposed framework consists of two main modules, the label prediction (LP) part and the common representation learning (CRL) part. \\ \\
\subsection{Notations}
Let the data from the two modalities $\mathcal{X},\mathcal{Y}$ be denoted as $D_x \in \mathbb{R}^{d_x \times N_x}$ and $D_y \in \mathbb{R}^{d_y \times N_y}$, where, $d_x$ and $d_y$ are the dimensions of the two modalities ($d_x \neq d_y$ in general).
The number of samples in both the modalities may be same i.e., $N_x=N_y=N$ or different i.e. $N_x \neq N_y$.
Out of them, $N^l$ data samples of both modalities are labeled, and the label matrix is denoted as $L \in \mathbb{R}^{d_c \times N^l}$, where $d_c$ is the total number of categories.
For single label (S-L) data, only one of the entries in $d_c$ is one, while for multi-label (M-L) data, multiple entries in $d_c$ can be one signifying the presence of multiple tags.
Thus, the number of unlabeled data in $\mathcal{X}$ and $\mathcal{Y}$ modalities are $N_x^{ul} = N_x - N^l$ and $N_y^{ul} = N_y - N^l$ respectively.
For ease of explanation, we will first consider the unlabeled data to be paired i.e., $N_x^{ul}=N_y^{ul}=N^{ul}$ and then extend it to handle the unpaired setting.
We denote the labeled and unlabeled portion of the training data as $D_x^{l}, D_y^{l}$ and $D_x^{ul}, D_y^{ul}$ respectively.

For performing cross-modal retrieval, first the data from the two modalities are projected to a common representation space, which is taken as the label space~\cite{gssl} \cite{s2cdl} in our work. 
In the proposed framework, two different deep networks are used to project the data of the two modalities.
To handle the unlabeled training samples, we also design a label prediction module which is trained to predict the labels for both S-L and M-L data.
The predicted labels are then fed to the main representation learning branch of the network.
This is inspired from~\cite{minsup}, though there are significant differences as explained in the next section.
The two parts of the network are trained sequentially one after the other (Figure \ref{main_flowchart}).
We now describe the two modules of the proposed framework in details below.

\subsection{Label Prediction (LP)}

We design a Label Prediction (LP) network to predict the labels (S-L or M-L) of the unlabeled portion of the training data of both the modalities. 
Our network is inspired by the label cleaning framework proposed in~\cite{minsup}, though in this work, they had access to the noisy labels and the goal was to clean the noisy labels for the task of single modality classification.
In contrast, a portion of the training data in our work is completely unlabeled and the task is cross-modal retrieval.

First, we generate weak labels/annotations of the unlabeled training data, i.e. $D_x^{ul}, D_y^{ul}$ by utilizing the labeled portion using a simple, but effective approach.
We randomly select a small portion of the labeled training data to form the Nearest Neighbor (NN) set as $\{D_x^{l,nn}, D_y^{l,nn}, L^{nn}\}$.
We also keep aside another portion as validation (val) set $\{D_x^{l,val}, D_y^{l,val}, L^{val}\}$ required for training the LP network.
The remaining labeled data $\{D_x^{l,tr}, D_y^{l,tr}, L^{tr}\}$ forms the train set for the LP network.
Considering the NN set to be a set of anchor points, we use them to compute the weak/noisy annotations for the val and train set. 
For each paired data sample $(x, y) \in ( D_x^{l,tr}, D_y^{l,tr} )$, we find its closest match in $( D_x^{l,nn},D_y^{l,nn} )$ and assign its label $( L_x^{n,tr}, L_y^{n,tr} )$ as the noisy label for the pair $(x, y)$. 
Since the training data is paired, we do a logical ``OR" ($f^{or}$) operation to generate the final noisy label $L^{n,tr} = f^{or}(L_x^{n,tr}, L_y^{n,tr})$.
The same approach is followed for generating the weak annotations for $( D_x^{l,val},D_y^{l,val} )$.

\begin{figure}[t!]
	\begin{center}		
		\includegraphics[width=10.5cm,height=6.5cm]{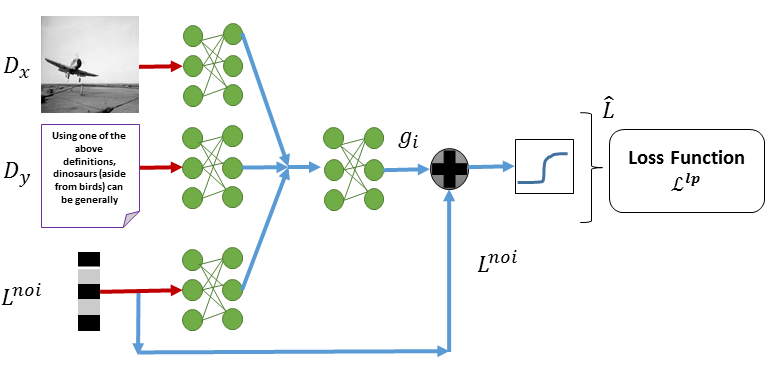}
	\end{center}
	\caption{ Illustration of the label prediction network. Here, the input to the network (with brown arrows) are image data, text data, noisy labels and the output are the predicted labels. 
	The activation function helps to scale the prediction between $[0,1]$ in case of M-L data. 
    For S-L data, the activation function is replaced with a softmax function i.e., $f^{soft}(.)$. 
The loss function used is $\mathcal{L}^{wbce}$ for M-L and $\mathcal{L}^{ce}$ for the S-L dataset respectively. }
	\label{label_prediction}
	\vspace{-10 pt}
\end{figure}


Thus, the problem of label prediction for the unlabeled data is transformed to a problem of label cleaning, where the  noisy labels are generated as described above.
The LP network consisting of fully connected (fc) layers and skip connection, takes the data from the two modalities $D_x^{l,tr}, D_y^{l,tr}$, their noisy labels $L^{n,tr}$ as input, and tries to predict the correct labels i.e., $L^{tr}$. 
We use both the modalities together to utilize the complementary information to better predict the missing annotations. An illustration of the LP network is given in Figure \ref{label_prediction}, where we first transform the three inputs using a fc layer and then concatenate the outputs before sending it through two additional fc layers. 
The output of the final fc layer for the $i^{th}$ data sample, $g_i$ has the same dimensionality as the label vector i.e., $d_c$ and is matched with the correct labels $L^{tr}_i$.
Instead of learning the complete clean label from scratch, we learn the residual and then use an identity connection \cite{resnet} \cite{minsup} from the input noisy label $L^{n,tr}_i$ to get to the correct label $L^{tr}_i$.

For M-L data, the predicted labels are computed as $\hat{L}^{tr}_i = \text{clip} (L^{n,tr}_i + g_i, [0,1]), $
where the clip operation forces the predicted label to be within the correct range i.e., $[0,1]$. We then compute the loss between the predicted $\hat{L}^{tr}_i = \{ \hat{L}^{tr}_{i,1}, \hat{L}^{tr}_{i,2}, ..., \hat{L}^{tr}_{i,d_c} \}$
and the correct $L^{tr}_i = \{ L^{tr}_{i,1}, L^{tr}_{i,2}, ..., L^{tr}_{i,d_c} \}$ and back-propagate the gradients to train the network. We use the weighted binary cross entropy (WBCE) loss as 
\begin{eqnarray}
	\label{wbce}
	\begin{aligned}
		\mathcal{L}^{wbce}(L^{tr}_i, \hat{L}^{tr}_i) = - \sum_{j=1}^{d_c} [ w^j L^{tr}_{i,j} \log ( \hat{L}^{tr}_{i,j} ) + (1-L^{tr}_{i,j}) \log ( 1 -  \hat{L}^{tr}_{i,j} ) ]
	\end{aligned}
\end{eqnarray}
where, the loss $\mathcal{L}(L^{tr}_i, \hat{L}^{tr}_i)$ is computed across each label vector dimension i.e., $j=\{1,...,d_c\}$, $L^{tr}_i, \hat{L}^{tr}_i$ are the two arguments to the loss and $w^j \geq 1$ denotes the positive weight for dimension $j$. 
For multi-label datasets, the absence of a tag is more common compared to it being present, and hence we want to penalize marking a present tag as absent considerably more than the opposite using this weight.
The positive weights $w^j$ can be set empirically by studying the distribution of ones and zeros in the labeled portion of the data.
We also tried using the standard BCE and the absolute distance error measure as in \cite{minsup} in the LP framework, but the WBCE loss gave significant improvement in performance compared to the others.
We have analyzed the importance of this loss in more details later.

For S-L data, the noisy input to the LP network i.e., $L^{n,tr}_i$ is one-hot encoding representation. 
We make corrections using the skip connection as before $\hat{L}^{tr}_i = f^{soft}(L^{n,tr}_i + g_i) $ where ($f^{soft}(.)$) is the softmax layer used to predict the final label.
In this case, we use cross entropy loss (CE) to back-propagate the error.
\begin{eqnarray}
	\label{ce}
	\begin{aligned}
		\mathcal{L}^{ce}(L^{tr}_i, \hat{L}^{tr}_i ) = - \sum_{j=1}^{d_c} L^{tr}_{i,j} \log ( \hat{L}^{tr}_{i,j} )
	\end{aligned}
\end{eqnarray}
Thus, the final label prediction loss is given as
\begin{eqnarray}
	\label{prediction_loss}
	\begin{aligned}
		\mathcal{L}^{lp}=
		\begin{cases}
			\sum_{tr} \mathcal{L}^{wbce}(L^{tr}_i, \hat{L}^{tr}_i), & \text{for M-L} \\ \\
			\sum_{tr} \mathcal{L}^{ce}(L^{tr}_i,  \hat{L}^{tr}_i), & \text{for S-L}
		\end{cases}
	\end{aligned}
\end{eqnarray}
We train our LP network using $\{D_x^{l,tr}, D_y^{l,tr}, L^{n,tr}\}$ and track its performance progress by computing the accuracy on the validation set $\{D_x^{l,val}, D_y^{l,val}, L^{n,val}\}$.


As we will see in the next section, for training the CRL network, the labels of the unlabeled training data predicted by the LP network are used.
One pertinent question here is how to feed the data to the LP network if the unlabeled training data is unpaired. 
In this case, for an unlabeled sample $x \in D_x^{ul}$ whose paired sample in $D_y^{ul}$ is unavailable, we compute its nearest neighbor in $D_x^{l,nn}$ and mark its weak annotation as $L_x^{n,ul}$. We find the closest match of $x$ in $D_x^{l,tr}$ and set its corresponding data sample in $\mathcal{Y}$ domain as $y$ and its label as $L_y^{n,ul}$. We then generate the noisy label $L^{n, ul}$ as previously done.


\subsection{Common Representation Learning (CRL)}

\begin{figure}[t!]
	\begin{center}		
		\includegraphics[width=12.5cm,height=8.0cm]{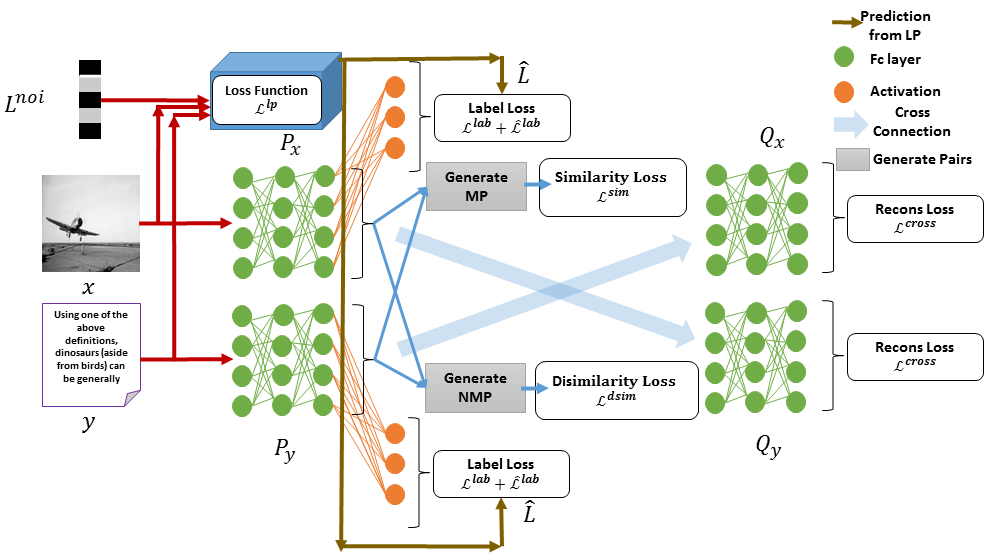}
	\end{center}
	\caption{Illustration of the proposed framework. The pre-trained LP block (marked in blue) outputs the predicted labels for the unlabeled data and feeds it to the label loss $\mathcal{\hat{L}}^{lab}$. 
	The LP network also takes as input the noisy labels for the unlabeled data, which can be generated both for paired and unpaired data. 
	The label loss for the labeled examples $\mathcal{L}^{lab}$ is computed from the output of the activation layer. 
The {\em generate pair} block generates matched and non-matched pairs from the two modalities which is used to compute the similarity $\mathcal{L}^{sim}$ and dissimilarity $\mathcal{L}^{dsim}$ losses respectively. 
Also, the outputs of the decoders are used to compute the cross reconstruction losses $\mathcal{L}^{cross}$. (Figure  best viewed in color).}
	\label{main_flowchart}
	\vspace{-10 pt}
\end{figure}

For cross-modal retrieval, we need to project the data from the two different modalities into a common space, where they have a shared representation.
In this work, based on the success of \cite{s2cdl} and \cite{gssl}, we select the label space i.e., $\mathbb{R}^{d_c}$ for this purpose.
For each modality, we use an encoder-decoder architecture with the the size of the bottleneck layer being equal to $d_c$.
An illustration showing our proposed architecture is given in Figure~\ref{main_flowchart}.
For learning the common representation, we project both the labeled $D_x^{l}, D_y^{l}$ and unlabeled $D_x^{ul}, D_y^{ul}$ data into the label space.
Ideally, we want the following (a) the projection of the data from the two modalities should be consistent with their labels, (b) similar data should lie close together while dissimilar data should be far apart and (c) the bottleneck code so generated should preserve most of the information from the original features.
To satisfy the above three objectives we design our loss functions appropriately.

Consider the neural network functions defined by the encoder and decoder for the $\mathcal{X},\mathcal{Y}$ modalities be denoted as $\mathcal{P}_x(.), \mathcal{P}_y(.)$ and $\mathcal{Q}_x(.), \mathcal{Q}_y(.)$. 
Both the encoder and decoder networks consists of three fc layers, with the decoder layers having mirrored sizes to that of the encoder. 
To make the projections of $\mathcal{P}_x(.), \mathcal{P}_y(.) \in \mathbb{R}^{d_c}$, the final activation function used is the sigmoid layer ($\sigma(.)$) for the M-L data and softmax layer ($f^{soft}(.)$) for the S-L data.
We now describe the different losses used in our work.  \\ \\ 
{\bf Label Loss:} The label loss $\mathcal{L}^{lab}$ tries to make the final encoder outputs consistent with that of the label space by minimizing the following objective (for $t=\{x,y\}$)
\begin{eqnarray}
	\label{label_loss}
	\begin{aligned}
		\mathcal{L}^{lab} =
		\begin{cases}
			\sum_{i=1}^{N_l} \sum_{t} \mathcal{L}^{wbce}(L_i, \sigma(\mathcal{P}_t(D_{t,i}^{l})) ) , & \text{for M-L} \\ \\
			\sum_{i=1}^{N_l} \sum_{t} \mathcal{L}^{ce}(L_i, f^{soft} (\mathcal{P}_t(D_{t,i}^{l})) ) , & \text{for S-L}
		\end{cases}
	\end{aligned}
\end{eqnarray}
{\bf Cross Reconstruction Loss:} The objective of the decoder is to reconstruct the input data from the common representation.
We enforce the reconstruction losses to reduce the amount of information loss from the main feature domain to the bottleneck layer.
In a standard encoder-decoder structure, the input to the decoders for the two modalities $\mathcal{Q}_x(.), \mathcal{Q}_y(.)$ should be the outputs of the corresponding encoders $\mathcal{P}_x(.), \mathcal{P}_y(.)$ respectively.
Here, since the labeled data is given in pairs (or pairs can be generated by studying the labels), we instead send the common representation generated by $\mathcal{P}_x(.)$ through $\mathcal{Q}_y(.)$ to reconstruct the data of the other modality and vice-versa.
This loss also helps to bridge the modality gap between the two inputs. 
We thus define our cross reconstruction loss as (for $t=\{x,y\}$ and $t'=\{y,x\}$)
\begin{eqnarray}
	\label{cycle_loss}
	\begin{aligned}
		\mathcal{L}^{cross} = \sum_{i=1}^{N_l} \sum_{t,t'} || D_{t,i}^{l} - \mathcal{Q}_t ( \mathcal{P}_{t'} (D^{l}_{t',i}) )||_1
	\end{aligned}
\end{eqnarray}
{\bf Similarity Dissimilarity Learning Loss:} The objective of this loss is to maintain the semantic structure of the data in the common representation. 
Though this is partly enforced by the label loss, here, we explicitly want to minimize the difference in the representations of all semantically similar data from the different modalities, while maximizing the distance between semantically different data. 
We observe that enforcing this at the outputs of the encoders $\mathcal{P}_x(.), \mathcal{P}_y(.)$ before the final activation function actually helps in the projection to the label space. 
We enforce this loss only on the labeled portion of the data i.e., $D_x^{l}, D_y^{l}$.
The semantically similar $(\mathcal{S}_1)$ and dissimilar sets $(\mathcal{S}_2)$ can be constructed from the provided labels $L$. 
For S-L data, we assume two data samples to be semantically similar if their labels are same, else they are dissimilar. 
For M-L data, we measure the similarity using normalized inner product $\mathcal(f^{s}(.)=<.,.>)$ between the labels.
If the similarity is more than a threshold $\tau_1$, we consider the samples as semantically similar, and if it is less than a threshold $\tau_2$ ($\tau_1 > \tau_2$), we consider them as semantically dissimilar.
Here we consider only the samples which are strongly similar or dissimilar for training, and we choose to ignore the small number of examples falling between the thresholds for which we are less confident.
Formally, two data samples are considered similar or dissimilar based on the following 
\begin{eqnarray}
	\label{sim_criterion}
	\begin{aligned}
		D_{x,i}^{l}, D_{y,j}^{l} \in \mathcal{S}_1 \hspace{2 pt} \text{if}, L_i=L_j \hspace{2 pt} \text{(S-L)} \hspace{2 pt} \text{or } f^{s}(L_i,L_j) \geq \tau_1 \hspace{2 pt} \text{(M-L)} \\
		D_{x,i}^{l}, D_{y,j}^{l} \in \mathcal{S}_2 \hspace{2 pt} \text{if}, L_i \neq L_j \hspace{2 pt} \text{(S-L)} \hspace{2 pt} \text{or } f^{s}(L_i,L_j) \leq \tau_2 \hspace{2 pt} \text{(M-L)} \nonumber
	\end{aligned}
\end{eqnarray}
We finally define the similarity and dissimilarity losses over the sets $\mathcal{S}_1$ and $\mathcal{S}_2$ respectively as
\begin{eqnarray}
	\label{sim_loss}
	\begin{aligned}
		\mathcal{L}^{sim} = \sum_{ (x_i, y_j) \in \mathcal{S}_1} || \mathcal{P}_x(x_i) - \mathcal{P}_y(y_j) ||_2^2
	\end{aligned}
\end{eqnarray}
\vspace{-20 pt}
\begin{eqnarray}
	\label{dsim_loss}
	\begin{aligned}
		\mathcal{L}^{dsim} = \sum_{ (x_i, y_j) \in \mathcal{S}_2} \max \left( 0, \mu - ||  \mathcal{P}_x(x_i) - \mathcal{P}_y(y_j) ||_2^2 \right)
	\end{aligned}
\end{eqnarray}
where, $x_i \in D_{x}^l, y_j \in D_{y}^l$ and $\mu$ is a margin set by cross-validation.
This implies that the distance between the representations of a non-matched or dissimilar pair must at least be $\mu$ distance apart, whereas the distance between matched pairs should be as small as possible. \\ \\
{\bf Total Loss:} Thus the total loss for the CRL network is given as 
\begin{eqnarray}
	\label{total_loss}
	\begin{aligned}
		\mathcal{L}^{crl} &=& \alpha_1 \mathcal{L}^{lab} + \alpha_2 \mathcal{L}^{cross} + \alpha_3 \mathcal{L}^{sim} + \alpha_4 \mathcal{L}^{dsim} + \beta \mathcal{\hat{L}}^{lab}
	\end{aligned}
\end{eqnarray}
where, $\mathcal{\hat{L}}^{lab}$ is the label loss with respect to the predicted labels provided by pre-trained LP for the unlabeled data $D_x^{ul}, D_y^{ul}$.
The variables $\{ \alpha_1, \alpha_2, \alpha_3, \alpha_4, \beta \}$ shows the different weights of the losses and is set by cross-validation experiments.


\subsection{Model training}
Here, we briefly discuss the procedure to train the two parts of the network.
The LP network in the proposed framework is trained initially based on the $\mathcal{L}^{lp}$ loss whose training is stopped based on the accuracy on the validation set $\{D_x^{l,val}, D_y^{l,val}, L^{n,val}\}$.
The loss function for training the CRL network (\ref{total_loss}) however deals with both the labeled and unlabeled portion of the data.
The unlabeled wbce loss $\mathcal{\hat{L}}^{lab}$ is based on the predicted labels of the pre-trained LP network.
In this work we have trained the two deep sub-networks separately because studies in \cite{minsup} showed no significant performance improvement on joint training of the two networks.
During testing, the common representation of the query and the data in the retrieval set are computed, which are used for retrieving items similar to the query.



\begin{table*}[]
	\small
	\centering
	\renewcommand{\arraystretch}{1.30}
	\setlength{\tabcolsep}{1.6 pt}
	\caption{Performance comparison (MAP@R scores at R=50 and R=all) on the S-L Wiki \cite{wiki} and M-L Pascal VOC \cite{pascal} and NUS-WIDE \cite{nus} datasets. The retrieval for the image query and text query are denoted as I-Q and T-Q respectively. The average results marked as Avg. is also reported. We have denoted the modes in which the methods are operating as - ``s" (supervised), ``us" (unsupervised) and ``ss" (semi-supervised). The superscript ``p" and ``up" denotes whether the unlabeled training data is provided with paired correspondence or not. The algorithms in the ``ss" mode is evaluated following the split as in \cite{gssl}.}
	\label{table_all}
	\begin{tabular}{|c|ccc|ccc|ccc|ccc|ccc|ccc|}
		\hline
		Dataset & \multicolumn{6}{c}{Wiki \cite{wiki}} & \multicolumn{6}{c}{Pascal VOC \cite{pascal}} & \multicolumn{6}{c|}{NUS-WIDE \cite{nus}} \\ \hline
		Tasks & \multicolumn{3}{c}{R=50} & \multicolumn{3}{c|}{R=all} & \multicolumn{3}{c}{R=50} & \multicolumn{3}{c|}{R=all} & \multicolumn{3}{c}{R=50} & \multicolumn{3}{c|}{R=all} \\ \hline
		Methods & T-Q & I-Q & Avg. & T-Q & I-Q & Avg. & T-Q & I-Q & Avg. & T-Q & I-Q & Avg. & T-Q & I-Q & Avg. & T-Q & I-Q & Avg. \\ \hline
		CCA$_{us}$ & 0.312 & 0.285 & 0.299 & 0.187 & 0.216 & 0.201 & 0.411 & 0.395 & 0.403 & 0.294 & 0.307 & 0.300 & 0.321 & 0.331 & 0.3265 & 0.266 & 0.286 & 0.276 \\
		SCM$_s$ & 0.355 & 0.301 & 0.328 & 0.233 & 0.275 & 0.254 & - & - & - & - & - & - & - & - & - & - & - & - \\
		LCFS$_s$ & 0.368 & 0.271 & 0.319 & 0.204 & 0.271 & 0.237 & 0.450 & 0.486 & 0.468 & 0.335 & 0.427 & 0.381 & 0.545 & 0.674 & 0.610 & 0.336 & 0.474 & 0.405 \\
		MvDA$_s$ & 0.391 & 0.309 & 0.350 & 0.231 & 0.297 & 0.264 & - & - & - & - & - & - & - & - & - & - & - & - \\
		LGCFL$_s$ & 0.495 & 0.386 & 0.441 & 0.316 & 0.377 & 0.346 & 0.502 & 0.529 & 0.515 & 0.344 & 0.436 & 0.390 & 0.598 & 0.590 & 0.594 & 0.390 & 0.497 & 0.444 \\
		ml-CCA$_s$ & 0.452 & 0.368 & 0.410 & 0.287 & 0.352 & 0.312 & 0.609 & 0.520 & 0.564 & 0.388 & 0.430 & 0.409 & 0.647 & 0.568 & 0.607 & 0.390 & 0.468 & 0.429 \\
		GMLDA$_s$ & 0.469 & 0.327 & 0.398 & 0.288 & 0.315 & 0.302 & - & - & - & - & - & - & - & - & - & - & - & - \\
		GMMFA$_s$ & 0.475 & 0.331 & 0.403 & 0.296 & 0.315 & 0.306 & - & - & - & - & - & - & - & - & - & - & - & - \\
		GSS-SL$_s$ & 0.531 & 0.412 & 0.471 & 0.338 & 0.406 & 0.372 & 0.655 & 0.556 & 0.605 & 0.412 & 0.466 & 0.439 & 0.682 & 0.869 & 0.775 & 0.405 & 0.550 & 0.477 \\
		\textbf{Ours$_s$} & \textbf{0.541} & \textbf{0.418} & \textbf{0.480} & \textbf{0.341} & \textbf{0.436} & \textbf{0.388} & \textbf{0.685} & \textbf{0.614} & \textbf{0.649} & \textbf{0.462} & \textbf{0.542} & \textbf{0.502} & \textbf{0.641} & \textbf{0.850} & \textbf{0.745} & \textbf{0.422} & \textbf{0.556} & \textbf{0.489} \\ \hline
		GSS-SL$_{ss}^{up}$ & 0.519 & 0.396 & 0.458 & 0.326 & 0.389 & 0.358 & 0.612 & 0.538 & 0.575 & 0.397 & 0.449 & 0.423 & 0.682 & 0.835 & 0.758 & 0.404 & 0.536 & 0.470 \\
		\textbf{Ours$_{ss}^{up}$} & \textbf{0.556} & \textbf{0.402} & \textbf{0.479} & \textbf{0.340} & \textbf{0.418} & \textbf{0.379} & \textbf{0.663} & \textbf{0.593} & \textbf{0.628} & \textbf{0.445} & \textbf{0.542} & \textbf{0.493} & \textbf{0.641} & \textbf{0.852} & \textbf{0.747} & \textbf{0.416} & \textbf{0.546} & \textbf{0.481} \\
		\textbf{Ours$_{ss}^{p}$} & \textbf{0.553} & \textbf{0.404} & \textbf{0.478} & \textbf{0.344} & \textbf{0.421} & \textbf{0.382} & \textbf{0.671} & \textbf{0.613} & \textbf{0.642} & \textbf{0.455} & \textbf{0.560} & \textbf{0.508} & \textbf{0.654} & \textbf{0.847} & \textbf{0.751} & \textbf{0.419} & \textbf{0.546} & \textbf{0.482} \\ \hline
	\end{tabular}
	\vspace{- 8 pt}
\end{table*}

\section{Experiments}

Here, we report the results of the proposed framework and compare it with the state-of-the-art approaches.
Specifically, we evaluate on three benchmark datasets, Wiki \cite{wiki} which is single-label and Pascal VOC 2007 \cite{pascal} and NUS-WIDE \cite{nus}, which are annotated with multiple labels for both supervised and semi-supervised settings.
Though all the experiments are on image-text retrieval, this method can work on any cross-modal retrieval application. Finally, we perform extensive analysis with respect to the amount of labeled data available and the loss functions used.

\subsection{Datasets and Evaluation Protocol}

The \textbf{Wiki dataset} \cite{wiki} contains about $2,866$ articles with their corresponding images and textual documents spread across $10$ different categories such as art, history, etc. collected from the Wikipedia repository.
We consider $4096$-d CNN \cite{caffe} feature representation for images and $100$-d word vectors \cite{word2vec} for texts. The train:test split is taken to be $2000:866$ out of which $1500$ samples of the training data are assumed to be labeled \cite{gssl}.

The \textbf{Pascal VOC 2007 dataset} \cite{pascal} consists of images and their textual queries annotated with multiple tags and the standard train:test split is $5011:4952$ images-text pairs.
As in \cite{gssl}, we remove all the pairs whose textual features are entirely zeros, thus giving the final train:test split as $5000:4000$.
We consider the labeled and unlabeled split to be $4000:1000$ and use 399-d word frequency features for text representation and $512$-d GIST features to describe the images as in~\cite{gssl}. 

The \textbf{NUS-WIDE dataset} \cite{nus} is a large dataset which has images and the associated tags spread across $81$ unique categories. 
We follow the same protocol as in \cite{gssl} and consider the data that belongs to the top $10$ largest groups, with a train:test split of $40834:27159$.
We use $500$-d SIFT and $1000$-d word frequency vectors for image and text representation respectively and $35000$ samples from the training data are considered to be labeled as in \cite{gssl}.

We use Mean Average Precision (MAP) as the evaluation metric which is defined as the mean of the average precision (AP) for all queries. 
AP can be defined as $AP(q) = \frac{\sum_{r=1}^{R}P_q(r) \delta(r)}{\sum_{r=1}^{R} \delta(r)}$, where $q$ is the query, $R$ is the number of retrieved items and $P_q(r)$ is the precision for query $q$ at position $r$.
MAP@R essentially measures the retrieval accuracy when $R$ number of items from the database are being retrieved per query item.
We report both MAP@50 and MAP@all for our experiments \cite{gssl}.
For S-L dataset like Wiki \cite{wiki}, a retrieved sample is considered to be correct (i.e., $\delta(r)=1$) if it has the same class as the query.
For M-L datasets, if the retrieved sample shares at least one common concept with the query, it is assumed to be correct.
We have compared the proposed approach against several recent cross-modal retrieval algorithms such as CCA \cite{cca}, SCM \cite{wiki}, GMLDA and GMMFA \cite{gma}, LCFS \cite{wang}, MvDA \cite{mvda}, LGCFL \cite{kang}, ml-CCA \cite{mlcca} and GSS-SL \cite{gssl}.
In this setting, the train and test sets are disjoint, unlike that in cross-modal hashing, and so comparisons with hashing techniques have not been performed.

\subsection{Results under supervised settings}

First, we evaluate the proposed approach for supervised setting where all the training samples are annotated with labels. 
The results of the proposed framework denoted as $Ours_s$ for the three datasets are provided in Table \ref{table_all}.
The other numbers are directly taken from the work in \cite{gssl}.
We observe that the proposed approach outperforms the state-of-the-art for Wiki and Pascal VOC datasets and gives comparable performance for NUS-WIDE. 
We also observe that for R@all, the improvement obtained by the proposed approach over GSS1-SL is more which signifies that our performance degrades gracefully when larger number of items are retrieved.

\subsection{Results under semi-supervised settings}

\begin{figure}[t!]
	\centering
	\begin{minipage}{.95\textwidth}
		\includegraphics[width = 0.95\textwidth, height=6.5cm]{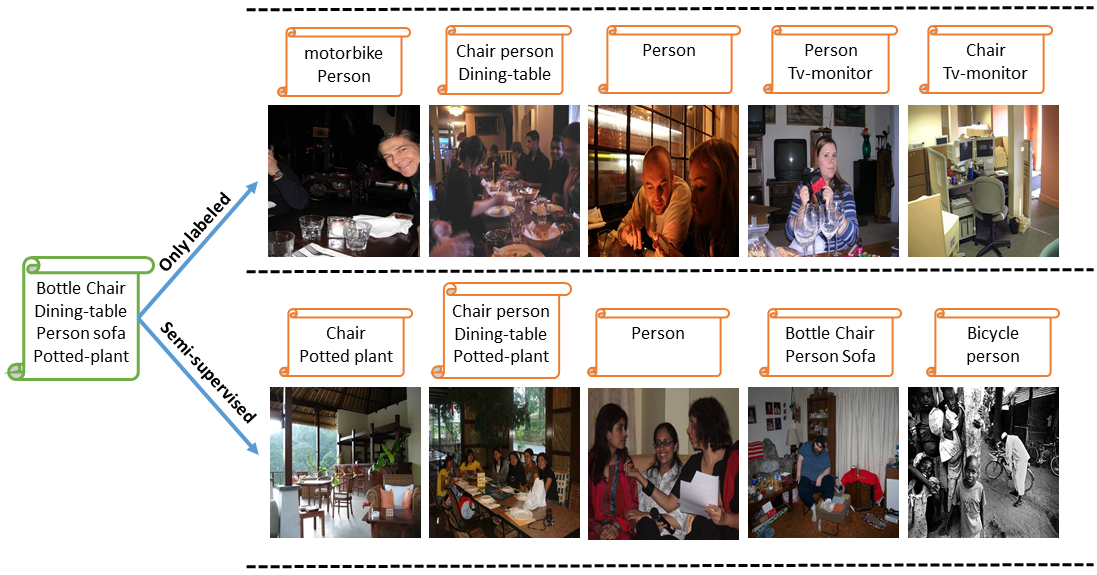}
	\end{minipage}%
	\hspace{2 pt}
	\begin{minipage}{0.95\textwidth}
		\includegraphics[width= 0.95\textwidth,height=6.5cm]{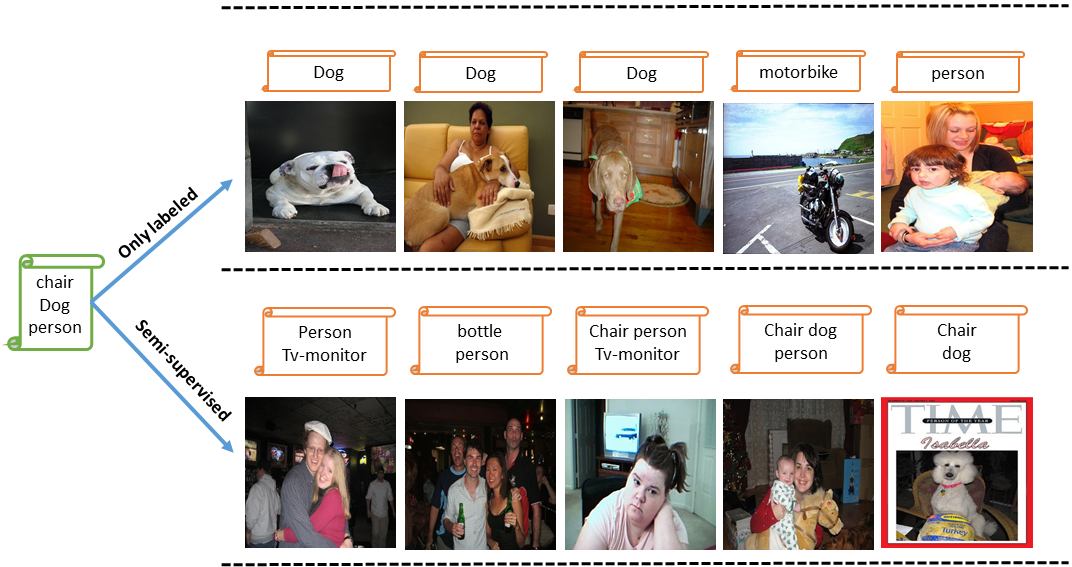}
	\end{minipage}
	\caption{Some text-image retrieval results for Pascal VOC \cite{pascal} using our algorithm in semi-supervised mode with $20\%$ labeled data. The actual tags are shown. We observe a noticeable improvement in the retrieval performance using the additional unlabeled data.}
	\label{improved}
	\vspace{-10 pt}	
\end{figure}

Here, we evaluate the proposed framework under the semi-supervised setting on the three datasets by taking the same split as in \cite{gssl}. 
Our results are denoted as $Ours^{p}_{ss}$ and $Ours^{up}_{ss}$ for the paired and unpaired scenarios respectively in Table \ref{table_all}.
We observe that even for the semi-supervised scenario, the proposed framework outperforms GSS1-SL.
As expected, the performance is slightly lower than the supervised counterpart.
Another interesting observation is that the paired setting usually gives slightly better performance.
This signifies the importance of the complementary information that is obtained from the correspondence information of the two modalities. 
Some retrieval results for the Pascal VOC dataset is shown in Figure \ref{improved}. 
We observe that the unlabeled data helps to retrieve more meaningful images from the database. 
Notice that a completely irrelevant image of ``motorbike" has been retrieved in the supervised case which has been corrected in the semi-supervised mode.

We also evaluate the importance of each loss in our formulation on Wiki \cite{wiki} and Pascal VOC \cite{pascal} datasets for $20\%$ labeled data. 
We observe from Table \ref{diff_loss_formula} that each of the losses contributes to the good performance of the proposed framework, though in varying amounts.

\subsection{Effect of varying amount of labeled data}

Here, we perform an experiment to analyze two things, (1) performance of the proposed framework for varying amounts of labeled data and (2) usefulness of the unlabeled data in improving the retrieval performance, which also is a measure of the effectiveness of the LP framework.
To do this, we vary the percentage of labeled data starting from $20\%$, with increments of $10\%$ to $90\%$ and report the performance of the proposed framework.
The remaining data is unlabeled and paired for this experiment.
In the semi-supervised mode, we use the predictions of the LP network.

\begin{figure}[t!]
	\centering
	\begin{minipage}{.48\textwidth}
		\centering
		\includegraphics[width = 0.85\textwidth, height=5.0cm]{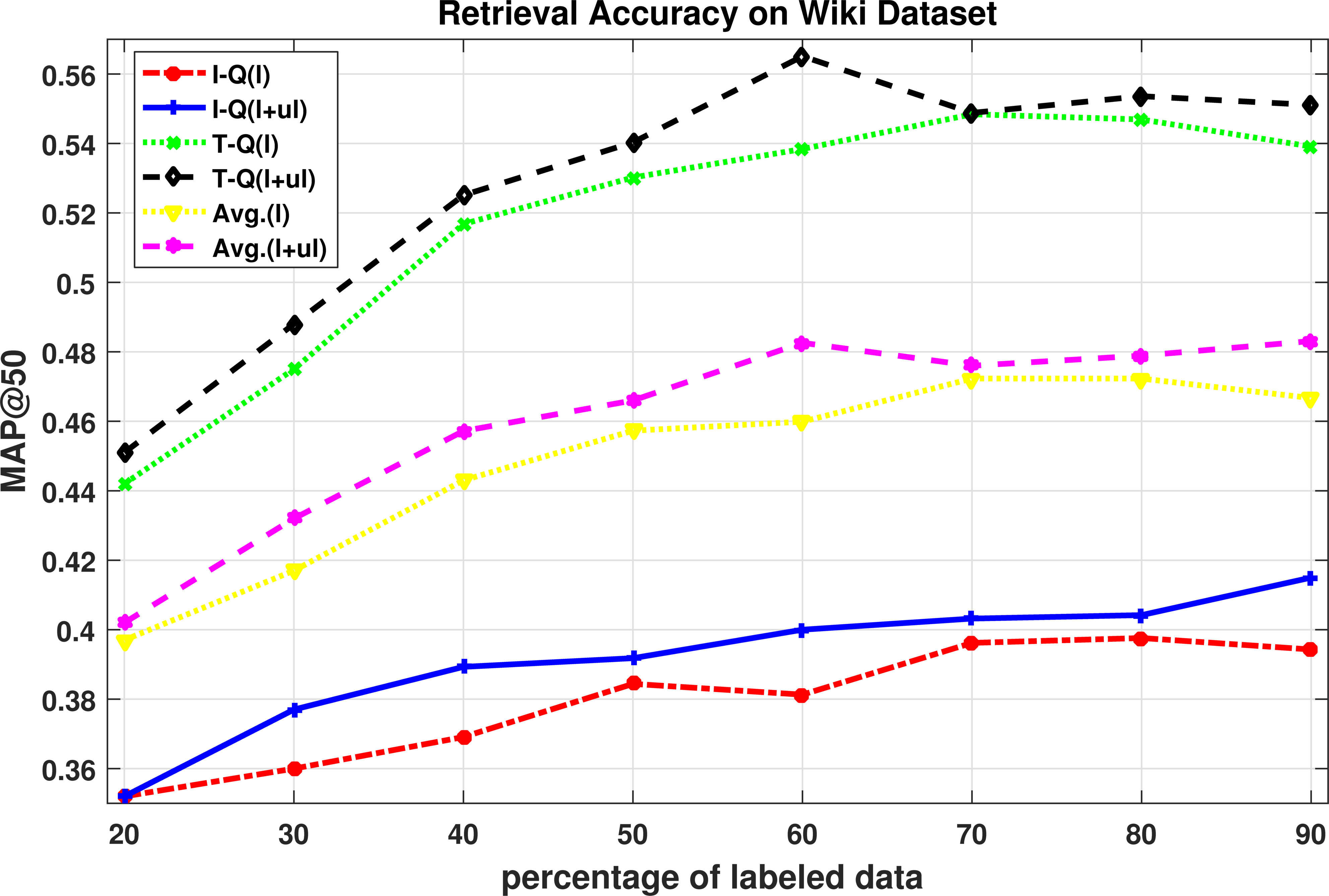}
		\caption{MAP@50 for Wiki~\cite{wiki} with increasing percentage of labeled data.}
		\label{wiki_r_50}
	\end{minipage}%
	\hspace{2 pt}	
	\begin{minipage}{.48\textwidth}
		\centering
		\includegraphics[width = 0.85\textwidth, height=5.0cm]{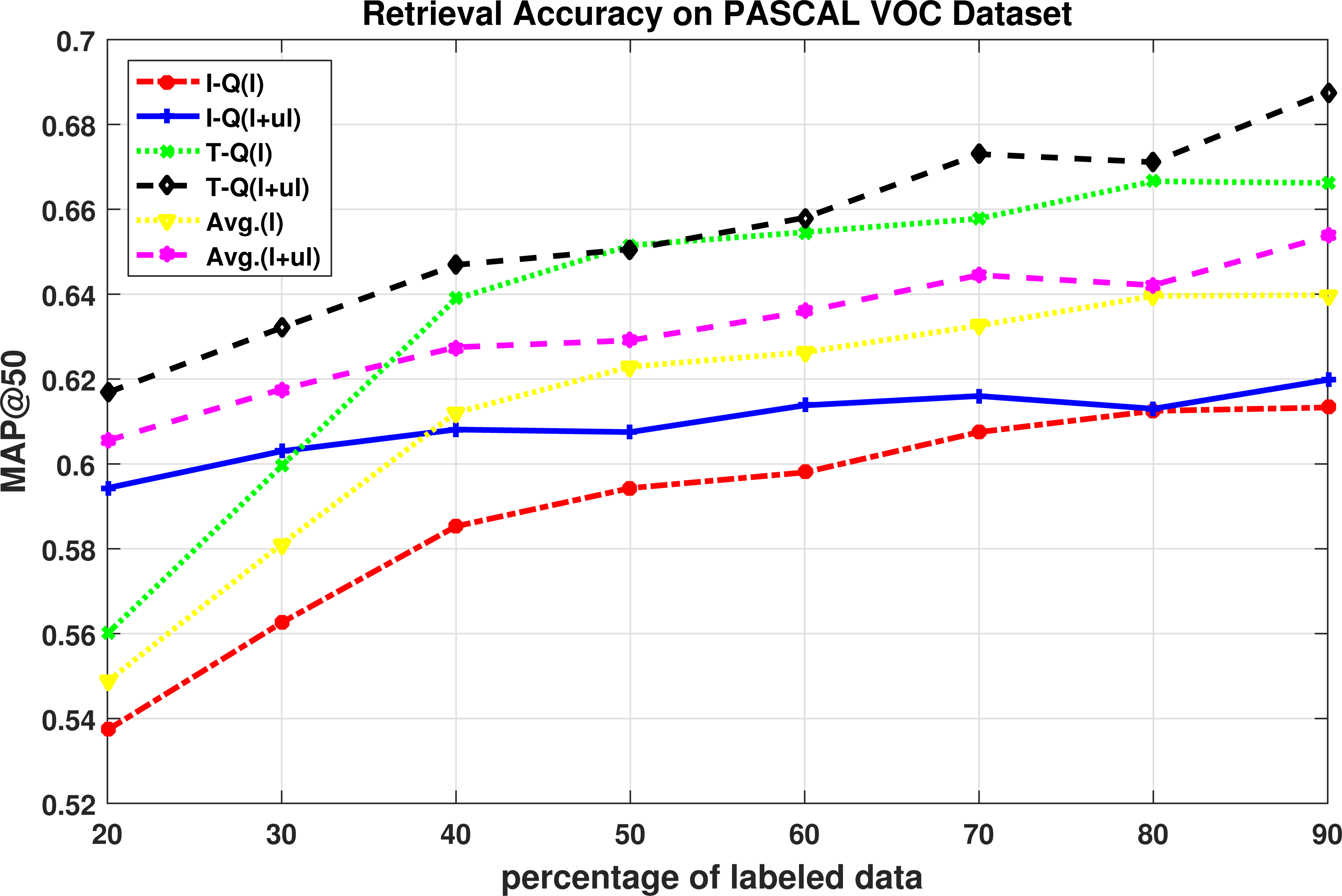}
		\caption{MAP@50 for Pascal VOC~\cite{pascal} with increasing percentage of labeled data.}
		\label{pascal_r_50}
	\end{minipage}%
\end{figure}

\begin{figure}[t!]
	\centering
	\begin{minipage}{.48\textwidth}
		\centering
		\includegraphics[width = 0.85\textwidth, height=5.0cm]{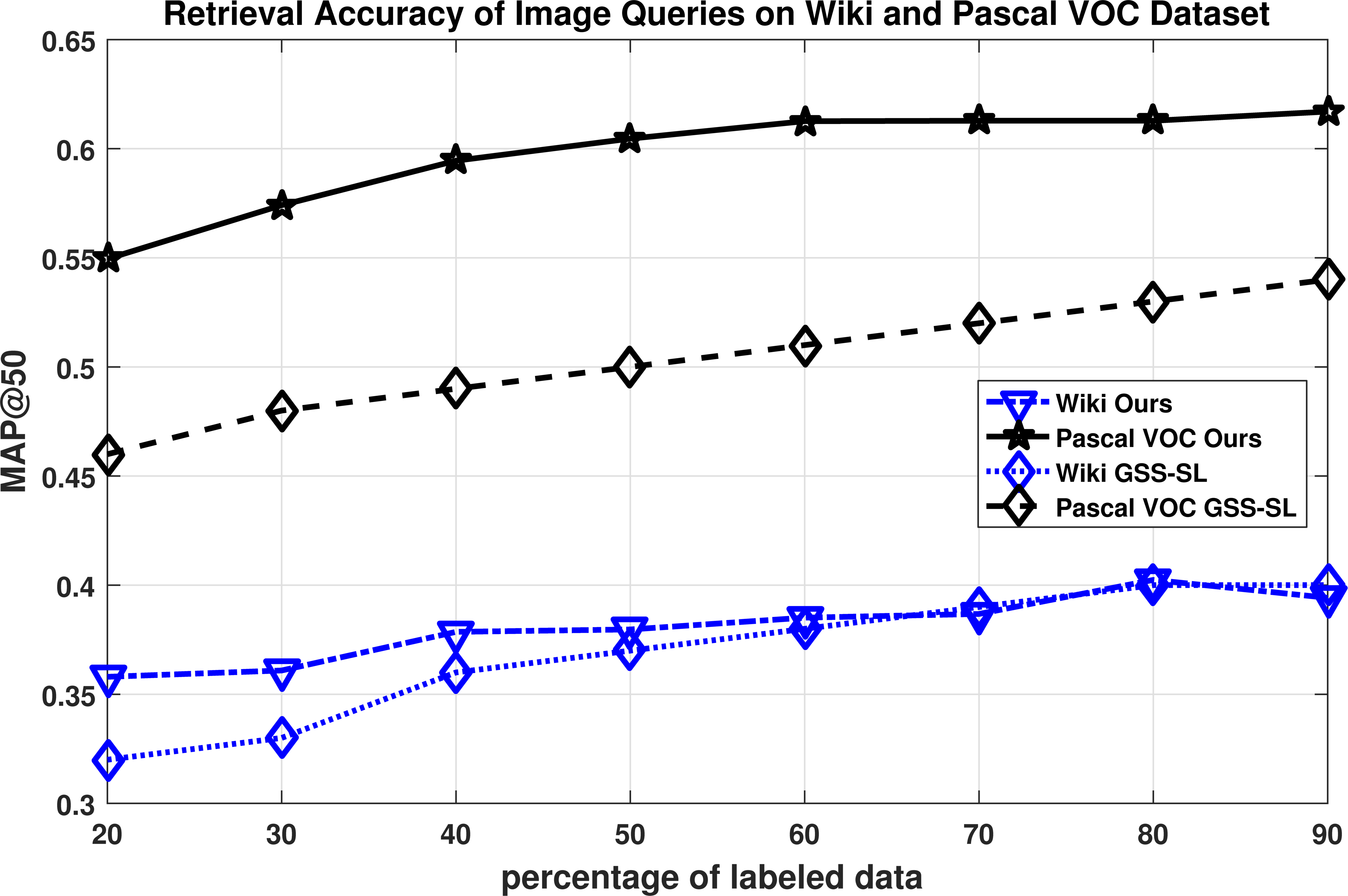}
		\caption{MAP@50 using $Ours^{up}_{ss}$ (a,b) and GSS-SL \cite{gssl} (c,d) on image queries for Wiki and Pascal VOC respectively with varying amounts of labeled data.}
		\label{image_up}
	\end{minipage}%
	\hspace{5 pt}
	\begin{minipage}{0.48\textwidth}
		\centering
		\includegraphics[width= 0.85\textwidth,height=5.0cm]{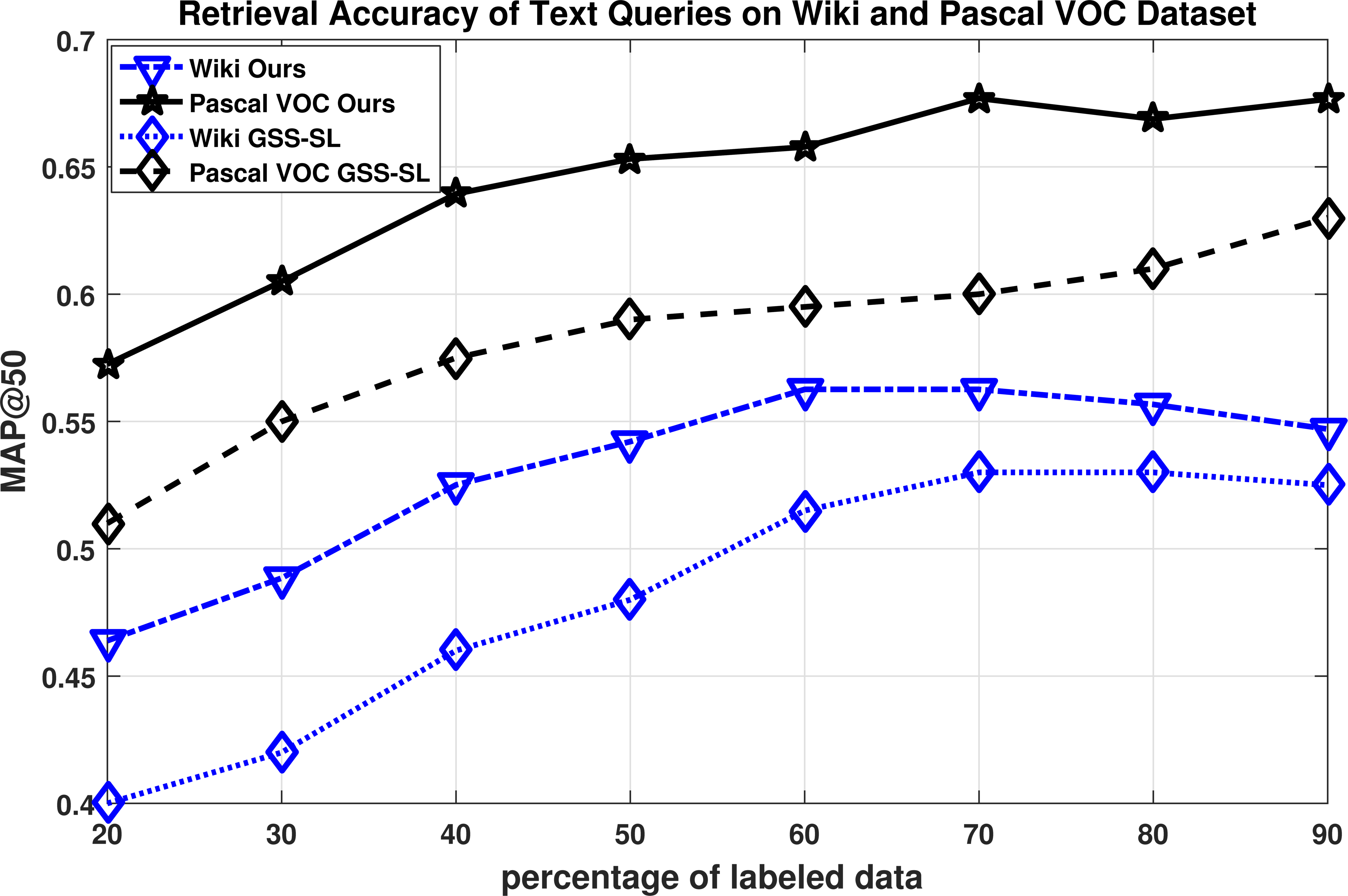}
		\caption{MAP@50 using $Ours^{up}_{ss}$ (a,b) and GSS-SL \cite{gssl} (c,d) on text queries for Wiki and Pascal VOC respectively with varying amounts of labeled data.}
		\label{text_up}
	\end{minipage}
\end{figure}

The MAP@50 results for the S-L Wiki \cite{wiki} dataset are reported in Figure~\ref{wiki_r_50}, and for the M-L Pascal VOC \cite{pascal} dataset are reported in Figure~\ref{pascal_r_50}.
We denote the results as I-Q (query is image), T-Q (query is text) and Avg (Average MAP). 
The superscript \emph{l} and \emph{ul} signifies that the algorithm is working in supervised and semi-supervised mode respectively.
We make the following observations from the results: (1) as expected, we notice that there is a monotonic increase in the retrieval percentage with the increase in the amount of labeled data; but after a certain stage it starts to saturate. This happens when the amount of labeled training data is sufficient to train the common domain projection functions i.e. $\mathcal{P}_x(.), \mathcal{P}_y(.)$; 
(2) we observe that the unlabeled data indeed helps to improve the performance, which also justifies the usefulness of the LP network; 
(3) when the amount of labeled data is less, the boost in performance provided by the unlabeled data is more as compared to when the labeled data is more. 
This is also partly because the amount of unlabeled data is more when there is less amount of labeled data. 
This is clearly evident from the results of Pascal VOC \cite{pascal} dataset.
We obtain similar observations when the unlabeled training data is also unpaired.
We also report the performance of $Ours^{up}_{ss}$ as compared to GSS1-SL \cite{gssl} in case of unpaired unlabeled data in Figures \ref{image_up} and \ref{text_up} for both image and textual queries. 
We observe similar trend as the paired setting, though the unpaired unlabeled data leads to less improvement in  performance compared to the paired unlabeled counterpart.

\begin{figure}[ht!]
	\centering
	\begin{minipage}{.48\textwidth}
		\centering
		\includegraphics[width = 0.85 \textwidth, height=5.0cm]{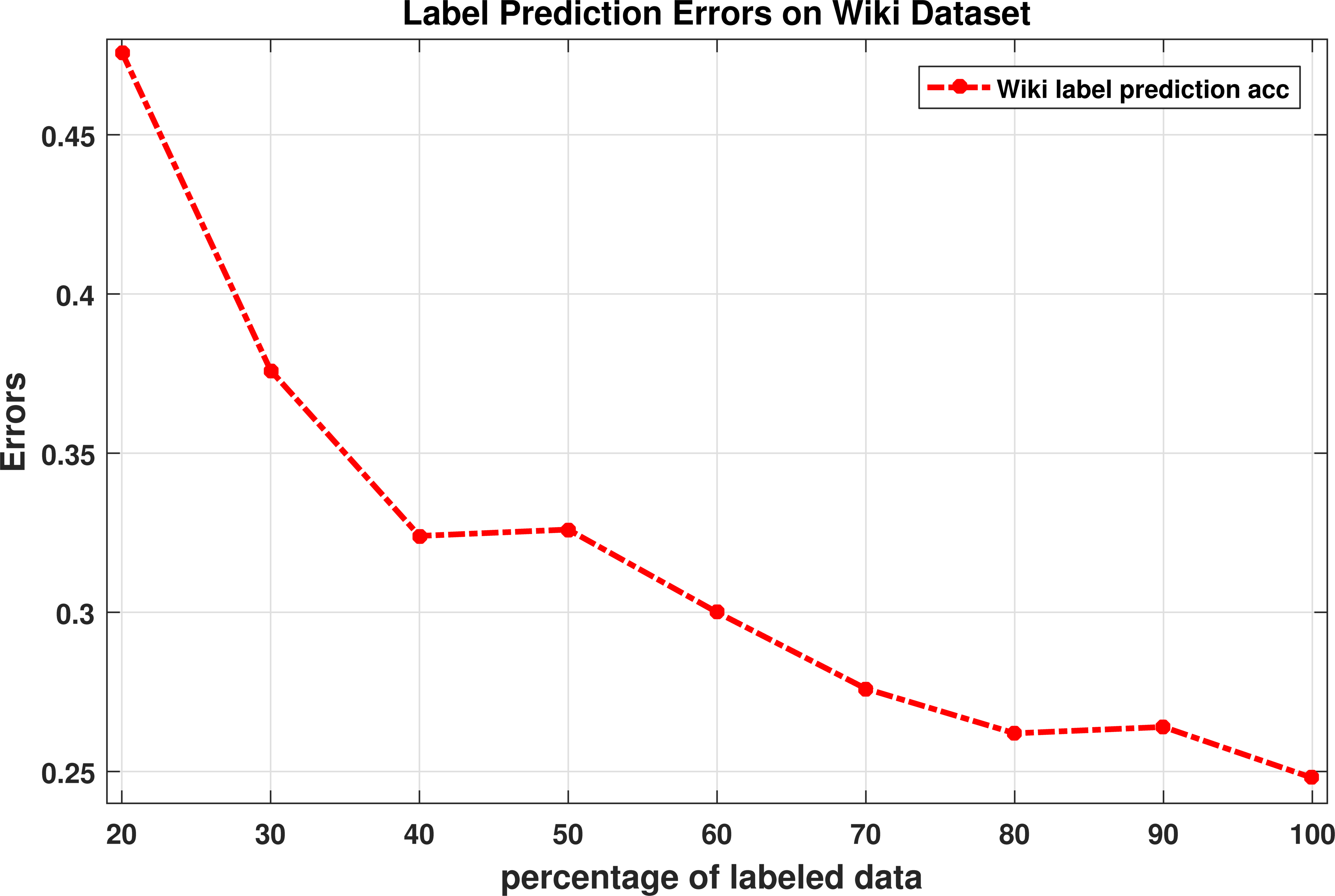}
		\caption{Prediction errors on Wiki.}
		\label{wiki_lp_err}
	\end{minipage}%
	\hspace{5 pt}
	\begin{minipage}{0.48\textwidth}
		\centering
		\includegraphics[width= 0.85 \textwidth,height=5.0cm]{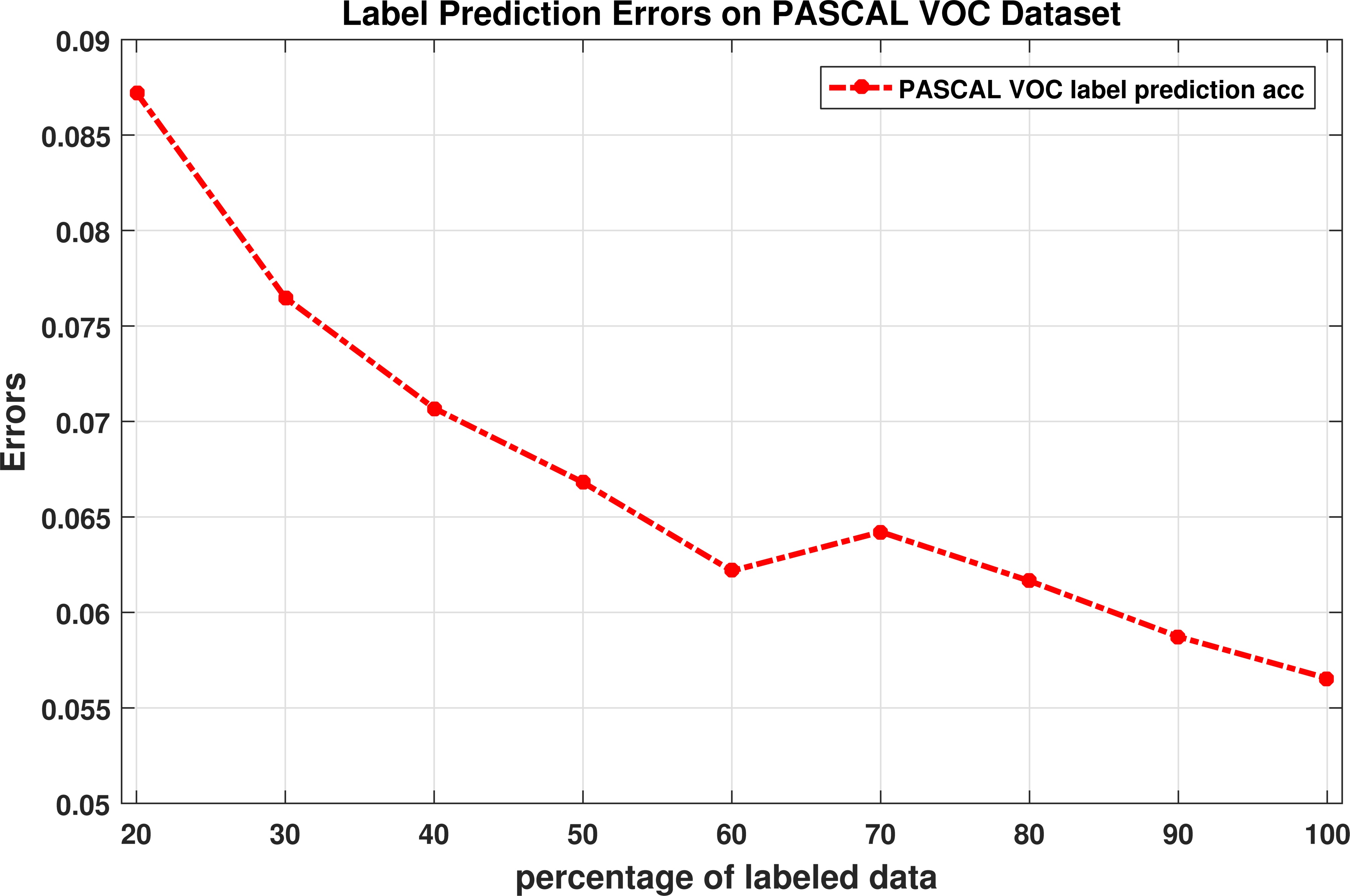}
		\caption{Prediction errors on Pascal VOC.}
		\label{pascal_lp_err}
	\end{minipage}
\end{figure}

We also analyze the performance of the label prediction network based on the amount of the labeled data. 
For S-L data, we consider a predicted label as correct if it matches with its ground truth. 
For M-L data, as previously discussed since the number of zeros is significantly higher than that of ones, we report the average number of errors in predicting ones and zeros as the accuracy. 
We measure the accuracy of the label prediction on the unlabeled paired data samples for the Wiki \cite{wiki} and Pascal VOC \cite{pascal} datasets, and the results as a function of amount of labeled data is given in Figures \ref{wiki_lp_err}, \ref{pascal_lp_err}.
As expected, greater the number of labeled examples, the easier it is to predict the labels for the unlabeled data. 

\begin{figure}[t!]
	\begin{center}		
		\includegraphics[width=0.42 \textwidth,height=5.0cm]{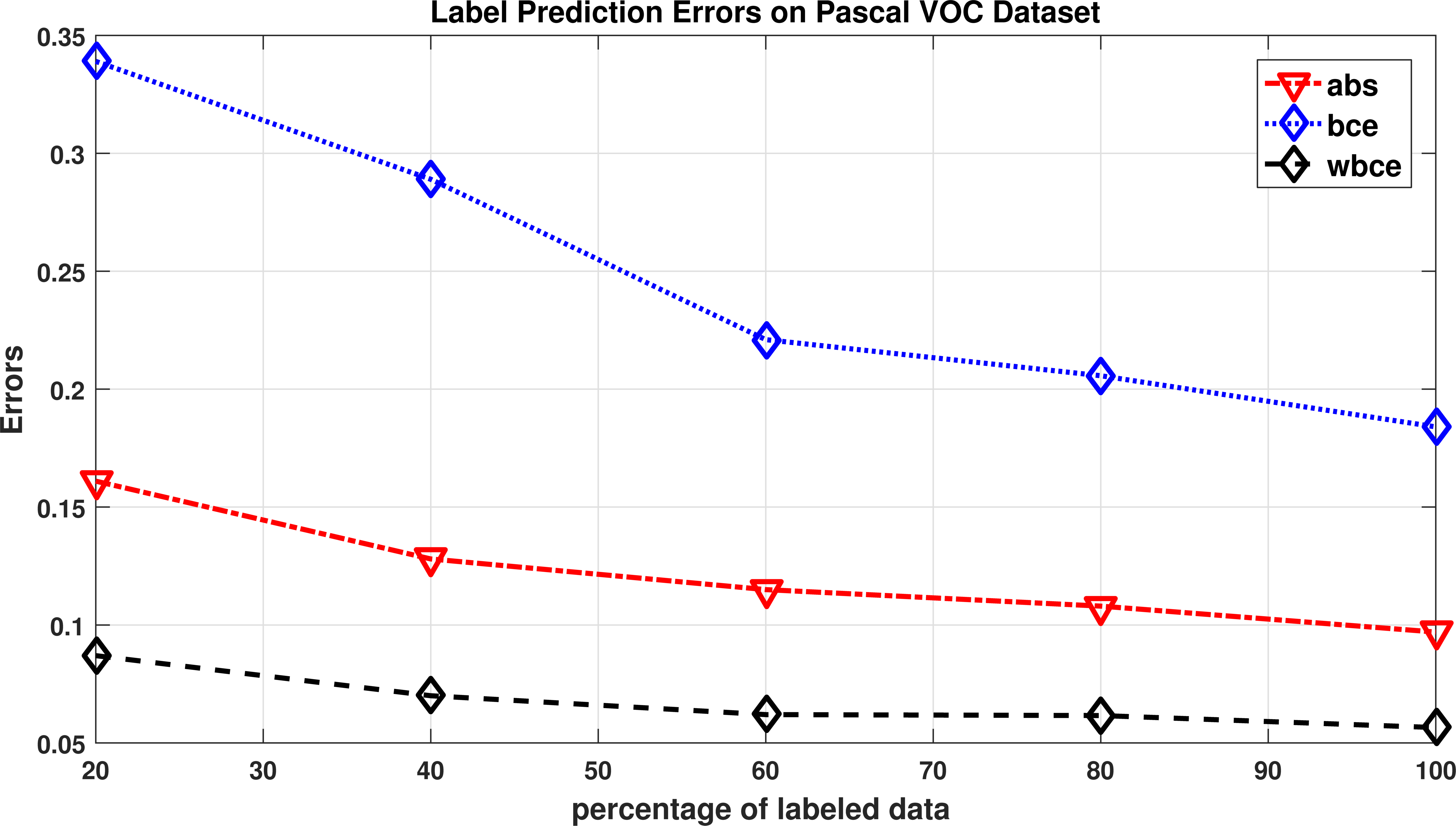}
	\end{center}
	\caption{Errors on Pascal VOC using different losses.}
	\label{voc_diff_losses}
\end{figure}

\subsection{Analysis of different losses for LP network}

We analyzed several standard losses for training the LP network. 
The CE loss was a natural choice for the S-L dataset.
For the M-L datasets, we analyzed the following loss functions (a) L1 loss (b) BCE and (d) WBCE.
The average prediction error of the LP network for the Pascal VOC dataset for the different losses is shown in Figure \ref{voc_diff_losses}.
We observe that the WBCE loss performs much better than the other losses, and we get the same conclusion for the 
other datasets also.
Thus, we finally selected WBCE for the M-L data for the label prediction task.


\begin{table}[h!]
	\small
	\centering
	\renewcommand{\arraystretch}{1.30}
	\setlength{\tabcolsep}{1.0 pt}
	\caption{MAP@50 and MAP@all for Wiki \cite{wiki} and Pascal VOC \cite{pascal} for $20\%$ labeled data with different losses being activated. The \checkmark signifies that all losses are present whereas the \text{\sffamily X} signifies that the particular loss is absent in (\ref{total_loss}).}
	\label{diff_loss_formula}
	\begin{tabular}{|c|c|cc|cc|}
		\hline
		\multicolumn{2}{|c|}{Dataset} & \multicolumn{2}{c|}{Wiki} & \multicolumn{2}{c|}{Pascal VOC} \\ \hline
		Losses & Presence & MAP@50 & MAP@all & MAP@50 & MAP@all \\ \hline
		$\mathcal{L}^{sim}$,$\mathcal{L}^{dsim}$ & \text{\sffamily X} & 0.392 & 0.303 & 0.569 & 0.464 \\ 
		$\mathcal{L}^{lab}$ & \text{\sffamily X} & 0.400 & 0.319 & 0.520 & 0.388 \\ 
		$\mathcal{\hat{L}}^{lab}$ & \text{\sffamily X} & 0.417 & 0.332 & 0.581 & 0.461 \\ 
		$\mathcal{L}^{cross}$ & \text{\sffamily X} & 0.430 & 0.353 & 0.618 & 0.480 \\ \hline
		$\mathcal{L}^{crl}$ & \checkmark & 0.433 & 0.357 & 0.618 & 0.481 \\ \hline
	\end{tabular}
\end{table}

\subsection{Implementation details}

We have implemented both of the networks LP and CRL in PyTorch. The LP network consist of a initial fc layer for the three inputs followed by a concatenation step and finally two additional fc layers. 
We have used fc layers of size $1,000$-d in all our experiments. 
The encoder-decoder architecture has three fc layers each of $5,000$-d. The hyper-parameters for the loss in~(\ref{total_loss}) i.e., $\{ \alpha_1, \alpha_2, \alpha_3, \alpha_4, \beta \}$ are selected to be $\{10, 1, 10, 1, 1\}$, $\{10, 1, 1, 1, 1\}$ and $\{10, 0.1, 0.1, 0.1, 10\}$ for the Wiki, Pascal VOC, and NUS-WIDE respectively. 
The learning rate $lr$ for the LP and CRL are set to be $\{0.005, 0.001\}$, $\{0.05,0.005\}$ and $\{0.005,0.0001\}$ for the Wiki, Pascal VOC, and NUS-WIDE datasets. 
We use stochastic gradient descent algorithm to train our networks.

\section{Conclusion}

In this work, we have proposed a deep-learning framework for the task of semi-supervised cross-modal retrieval.
This has been achieved by designing two sub-networks (a) a label prediction network for predicting labels of the unlabeled training data (both S-L and M-L) and (b) a common domain representation learning framework to project the data from different modalities into a common space. 
We propose different losses to learn an effective and discriminative common representation using both the labeled and unlabeled training data.
Extensive experiments performed on three standard benchmarking datasets have shown that the proposed framework outperforms the state-of-the-art for both semi-supervised as well as supervised settings. 
The proposed approach is able to effectively utilize the unlabeled data to give better retrieval performance, especially in the case of very small amounts of labeled data. 


\bibliographystyle{plainnat}
\bibliography{template}

\end{document}